\documentclass{article}
\usepackage[preprint]{neurips_2026}
\usepackage[utf8]{inputenc}
\usepackage[T1]{fontenc}
\usepackage{hyperref}
\usepackage{url}
\usepackage{booktabs}
\usepackage{amsfonts}
\usepackage{amsmath}
\usepackage{nicefrac}
\usepackage{microtype}
\usepackage{xcolor}
\usepackage{graphicx}
\usepackage{array}
\usepackage{caption}
\usepackage{adjustbox}

\title{OncoNoteBERT: A Foundation Representation Model for Natural Language Processing of Real-World Outpatient Oncology Notes}

\author{%
  Wuraola Oyewusi$^1$ \And
  Eliana Vasquez Osorio$^{1,3}$ \And
  Goran Nenadic$^2$ \And
  Gareth Price$^{1,3}$ \\
  $^1$ Division of Cancer Sciences, The University of Manchester, Manchester, UK \\
  $^2$ Department of Computer Science, The University of Manchester, Manchester, UK \\
  $^3$ The Christie NHS Foundation Trust, Manchester, UK \\
  \texttt{wuraola.oyewusi@postgrad.manchester.ac.uk} \\
  \texttt{\{eliana.vasquezosorio,gnenadic,gareth.price\}@manchester.ac.uk}
}

\begin{document}
\maketitle
\footnotetext{Accepted at the AI at Scale for Clinical Impact (ASCI): Cancer Pathology Foundation Models Workshop at NeurIPS 2026. \url{https://asci.artificialintelligencepathology.org/}}

\begin{abstract}
Real-world outpatient oncology notes contain specialised terminology, tumour staging expressions, treatment names, toxicity descriptions, and institution-specific de-identification markers that may not be represented efficiently by general biomedical or adjacent clinical language models. We developed and evaluated oncology-specific BERT-style encoders using a governed UK outpatient oncology corpus comprising 290,026 notes from 21,564 patients treated for lung and head-and-neck cancer. We compared RadBERT and PathologyBERT with two local strategies: OncoNote-RadBERT, produced by continued masked language model pretraining, and OncoNoteBERT, trained from scratch with an oncology WordPiece tokenizer. Models were evaluated using masked language modelling loss and perplexity on the validation set, together with tokenizer fragmentation metrics, clinical term tokenisation, masked-token probes, and exploratory representation analysis. We found that both external encoders fit the oncology corpus poorly in zero-shot evaluation (perplexity 113.04 for RadBERT, 2035.03 for PathologyBERT), while continued pretraining produced the strongest fit (perplexity 2.10 for OncoNote-RadBERT). OncoNoteBERT achieved perplexity of 2.83 but produced the most efficient tokenisation, with lower subword fertility and shorter normalised sequence length. It also returned a clinically acceptable prediction for 12 of 13 masked-token probes compared with 7 of 13 for OncoNote-RadBERT. This divergence between corpus-level language fit and masked-token performance was partly attributable to tokenizer fragmentation rather than learned semantics alone. Both locally developed models represented the institutional placeholder as a single learnable token, enabling explicit representation of redacted spans within the governed corpus. These findings show that continued adaptation and bespoke tokenisation provide complementary benefits, and that representation-layer design matters before adjacent-domain encoders are applied to oncology NLP.
\end{abstract}


\section{Introduction}

Natural language processing (NLP) makes clinical text accessible to computational analysis through numerical representations of tokens, sentences, and documents \citep{eisenstein2019nlp}. Clinical notes contain specialised vocabulary, frequent abbreviations, non-standard syntax, sentence fragments, and documentation structures that differ from general written text.This should be considered to avoid fragmenting important clinical concepts, making inefficient use of context-window capacity, and producing representations poorly aligned with the target domain.This is particularly relevant to transformer encoders such as BERT, which learn contextual representations through self-supervised masked language modelling and transfer them to downstream tasks \citep{liu2020representation,devlin2019bert}. For adapting such models to specialised clinical text, two broad strategies are continued pretraining of an existing model on target-domain text and training a new domain-specific tokenizer and model from scratch. Continued pretraining can exploit previously learned language representations while adapting model parameters to a target corpus \citep{gururangan2020dont}. From-scratch training can additionally align the tokenizer vocabulary with specialised terminology, abbreviations, and local documentation conventions \citep{gu2021domain}. Better intrinsic representations do not guarantee downstream gains but may provide a stronger foundation for task-specific modelling \citep{gutierrez2023biomedical}.

Several biomedical and clinical language models illustrate these approaches. BioBERT and ClinicalBERT adapted BERT-style models to biomedical literature and clinical notes, respectively \citep{lee2020biobert,alsentzer2019clinicalbert}, while RadBERT and PathologyBERT specialised representation learning for radiology and pathology text \citep{chambon2023radbert,santos2023pathologybert}. Oncology-specific BERT-based models have also been developed, for example CancerBERT, DFCI-ImagingBERT, MetBERT, OncoBERT, and Symptom-BERT \citep{zhou2022cancerbert,liu2022metbert,lin2023oncobert,zeinali2024symptombert,elmarakeby2023empirical}. However, these models were designed for downstream tasks rather than as general-purpose language encoders. Therefore, it remains unclear how best to build reusable representations for this setting, including whether continued masked language model pretraining or from-scratch model and tokenizer development provides a stronger foundation.

Moreover, oncology notes are longitudinal and multipurpose as they record diagnosis, staging, and imaging findings; treatments including surgery, systemic anticancer therapy, and radiotherapy; and patient preferences, supportive and complementary care, and treatment-related toxicities. This mixture creates a distinct language setting that may not be represented efficiently by adjacent-domain encoders. At the time of this study, we were not aware of any language model developed specifically from longitudinal outpatient oncology notes of this kind.

This study investigates how language representations for real-world UK outpatient oncology notes should be developed. We compare two domain-adjacent external encoders, RadBERT and PathologyBERT, with two oncology-adapted strategies: OncoNote-RadBERT, produced through continued Masked Language Modelling (MLM) pretraining from RadBERT, and OncoNoteBERT, trained from scratch using an oncology-specific tokenizer and randomly initialised BERT-base-style parameters. The work aims to develop and evaluate the intrinsic properties of general-purpose representation models for outpatient oncology language intended to provide a foundation for subsequent task-specific studies. We addressed three specific sub-questions: i) do domain-adjacent encoders trained on radiology or pathology text adequately represent outpatient oncology language? ii) does continued MLM pretraining on oncology notes improve intrinsic language fit? iii) does a bespoke oncology tokenizer represent clinical terminology more efficiently than an inherited vocabulary?

\section{Methodology}

\subsection{Study Design and Corpus}

\begin{figure}[t]
\centering
\includegraphics[width=0.92\linewidth]{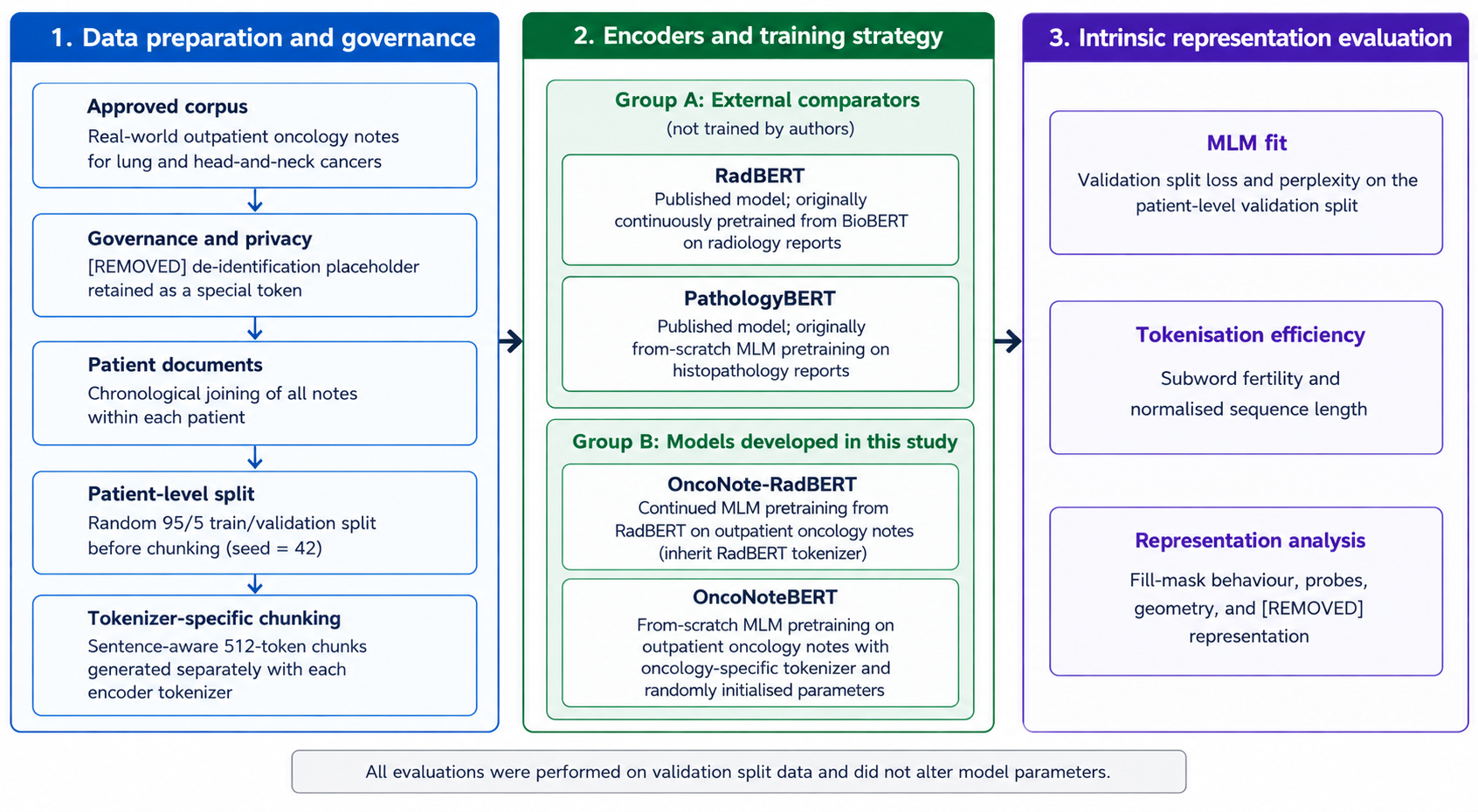}
\caption{Study workflow for data preparation, encoder development, and intrinsic representation evaluation. All evaluations were performed on the validation split of patient data and did not alter model parameters.}
\label{fig:1}
\end{figure}

We conducted a self-supervised representation-learning study using masked language modelling (MLM) on real-world outpatient oncology notes. The overall methodology is summarised in Figure \ref{fig:1}, which presents three stages: patient-level corpus preparation and governance, encoder selection and training strategy, and intrinsic representation evaluation.

\subsection{Data Preparation and Governance}

The study corpus comprised an approved subset of outpatient oncology notes from 21,564 patients treated for lung and head-and-neck cancer at a specialist oncology centre in the United Kingdom. The dataset contained 290,026 notes recorded from 11-July-2014 to 6-July-2026. All notes underwent institutional de-identification before use, with the placeholder \texttt{[REMOVED]} used to mark redacted information. This token was retained and represented as a dedicated token in the locally developed models.

For each patient, notes were ordered chronologically and concatenated to form a patient-level document. Whitespace was normalised, but stopwords were not removed, and no stemming, lemmatisation, placeholder deletion, or downstream task-specific cleaning was applied. Patients were randomly assigned to training and validation sets using a 95:5 patient-level split. Splitting was performed before tokenisation and chunking so that no text from the same patient could appear in both sets. The validation set comprised a separate patient subset used for intrinsic evaluation and checkpoint selection during training; the split was random rather than temporal.

\begin{table}[t]
\centering
\scriptsize
\caption{Patient-level training and validation corpus characteristics}
\label{tab:1}
\begin{adjustbox}{max width=\linewidth}
\begin{tabular}{@{}llllll@{}}
\toprule
Split & Patients & Notes & Median notes per patient & Median approximate tokens per patient document & \texttt{[REMOVED]} occurrences \\
\midrule
Training & 20,485 & 276,309 & 7 & 695 & 1,801,280 \\

Validation & 1,079 & 13,717 & 6 & 659 & 85,698 \\

Total & 21,564 & 290,026 & --- & --- & 1,886,978 \\
\bottomrule
\end{tabular}
\end{adjustbox}
\end{table}

\subsection{Encoders and Training Strategies}

\begin{figure}[htbp]
\centering
\includegraphics[width=0.82\linewidth]{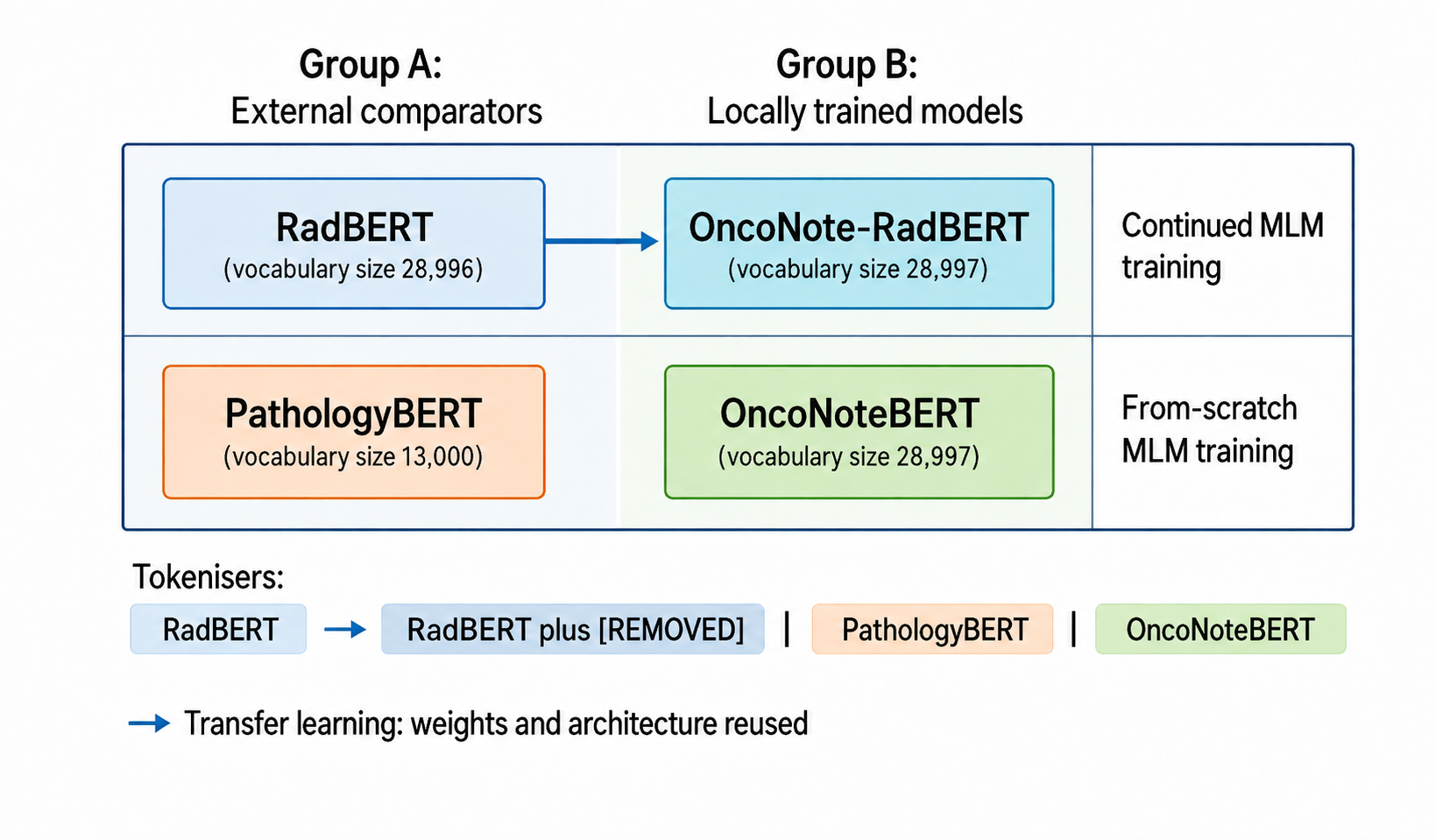}
\caption{Model groups, tokenizers, and training strategies}
\label{fig:2}
\end{figure}

The model groups, tokenizer choices, and training strategies are summarised in Figure \ref{fig:2}.

Group A - External comparators:

RadBERT \citep{chambon2023radbert} and PathologyBERT \citep{santos2023pathologybert} were selected as publicly available, domain-adjacent encoders representing two different development strategies. RadBERT was developed through continued pretraining from BioBERT on radiology reports and retained an inherited biomedical WordPiece tokenizer. PathologyBERT was trained from scratch on pathology reports using a pathology-specific tokenizer trained using WordPiece \citep{santos2023pathologybert}. Both encoders used a maximum input length of 512 tokens.

Group B - Locally trained models

OncoNote-RadBERT was initialised from the published RadBERT weights and tokenizer \citep{chambon2023radbert}. The \texttt{[REMOVED]} special token was added to the tokenizer vocabulary, and the embedding matrix was resized before continued MLM pretraining on the oncology corpus.

OncoNoteBERT used randomly initialised BERT-base-style parameters. A bespoke tokenizer was trained on our training dataset using WordPiece with a minimum token frequency of two, a continuation prefix of \#\#, and case sensitivity preserved. Its reserved vocabulary includes \texttt{[PAD]}, \texttt{[UNK]}, \texttt{[CLS]}, \texttt{[SEP]}, \texttt{[MASK]}, and \texttt{[REMOVED]}. Its final vocabulary size was 28,997 tokens, matching the RadBERT route after the addition of \texttt{[REMOVED]}.

OncoNote-RadBERT and OncoNoteBERT also used a maximum input length of 512 tokens. Because many patient-level documents exceeded this limit, each document was converted into tokenizer-specific chunks. Chunking was sentence-aware where possible and splitting within a sentence was used only when a single sentence exceeded the available token budget.

OncoNote-RadBERT and OncoNoteBERT were trained using dynamic masking of 15\% of eligible input tokens, per-device batch size of 32, gradient accumulation over two steps, an effective batch size of 64, weight decay of 0.01, a warm-up ratio of 0.06, and evaluation and checkpointing every 2,000 steps.

Training duration and early stopping were controlled by optimisation steps rather than fixed epochs as tokenizer-specific chunking produced different numbers of training sequences. OncoNote-RadBERT used a learning rate of 0.00005, a maximum of 100,000 steps, and patience of five evaluation rounds. OncoNoteBERT used a learning rate of 0.0001, a maximum of 200,000 steps, and patience of eight evaluation rounds. The longer schedule and patience accounted for the slower and less stable convergence expected from random initialisation.

Training was conducted within a dedicated data-resident environment to comply with data governance. Experiments used PyTorch 2.5.1 with CUDA 11.8 on a single NVIDIA RTX A6000 GPU with 47.99 GB of memory and bfloat16 precision. OncoNote-RadBERT and OncoNoteBERT were trained once, so the reported results should be treated as point estimates.

\subsection{Intrinsic Representation Evaluation}

The evaluation framework comprised held-out MLM fit, tokenisation efficiency, targeted clinical-term tokenisation, masked-token behaviour, and exploratory representation analysis. Following prior BERT-style MLM evaluation \citep{liu2019roberta}, perplexity was calculated as the exponential of the mean MLM validation loss. Tokenizer efficiency was assessed using subword fertility, continuation-token rate, unknown-token rate, and normalised sequence length relative to RadBERT. Qualitative MLM behaviour was assessed using 13 prespecified fill-mask prompts, recording whether a clinically acceptable answer appeared among the five highest-ranked predictions. These answers were prespecified using domain expertise and familiarity with the source oncology documentation, based on clinically plausible completions for each prompt. Finally, we calculated the mean cosine similarity across 750 matched validation chunks to estimate representation geometry. This analysis was treated as an exploratory audit rather than a performance endpoint because direct similarity is not a reliable measure of underlying representational similarity for independently trained embedding spaces \citep{kornblith2019similarity}.

\section{Results}

OncoNote-RadBERT training stopped after 88,000 steps, corresponding to approximately 51.4 passes through its chunked training dataset, which corresponded to 35.5h of training. OncoNoteBERT completed all 200,000 steps, corresponding to approximately 141.5 passes through its chunked training dataset, corresponding to 82.1h of training.

\subsection{Perplexity}

\begin{figure}[t]
\centering
\includegraphics[width=0.92\linewidth]{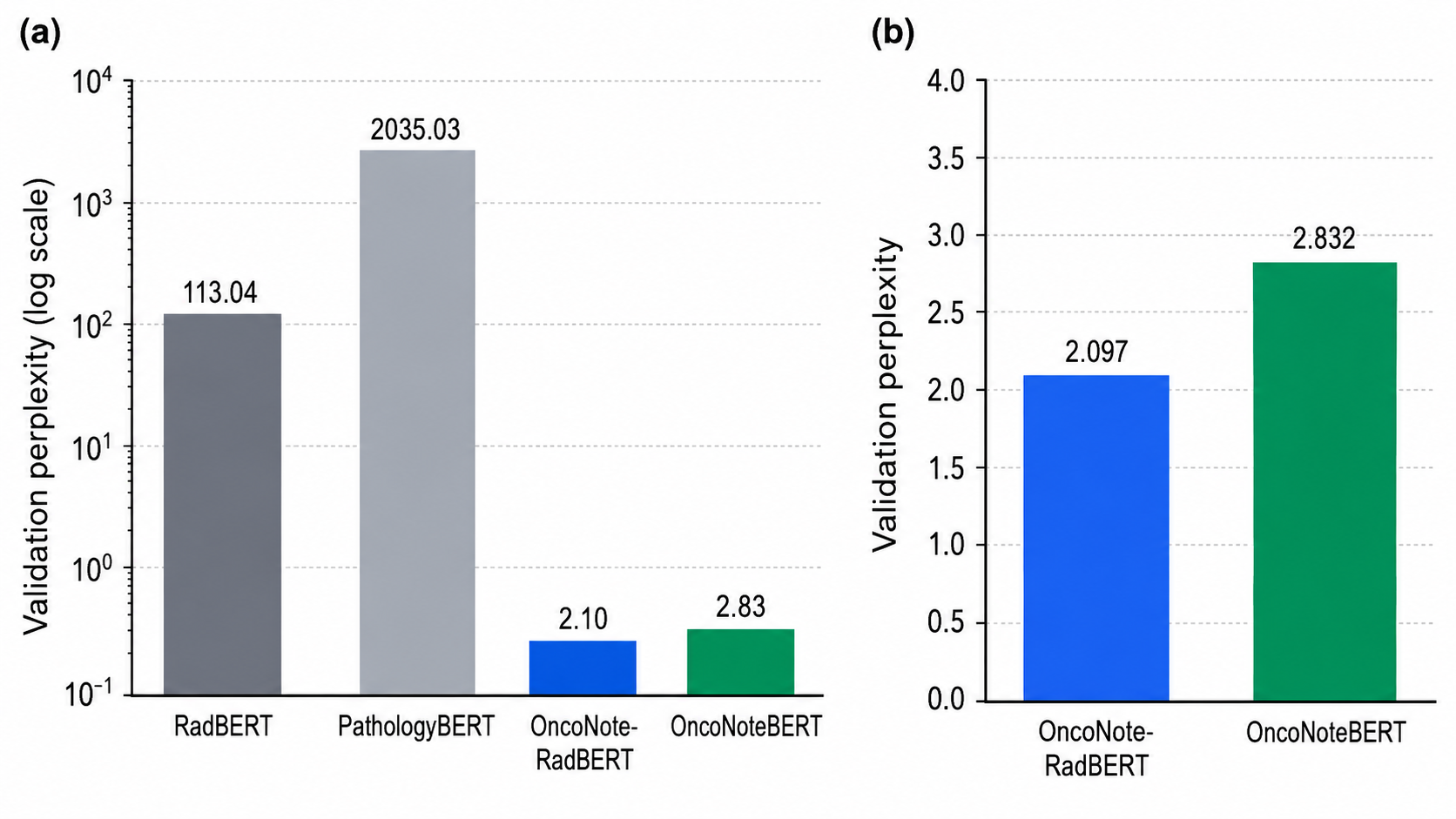}
\caption{Perplexity across the four encoders evaluated on the validation dataset. Lower perplexity indicates better language-model fit. Panel (a) shows the perplexity for all evaluated models on a log scale. Notice the substantially higher PathologyBERT perplexity. Panel (b) shows the perplexity of the two locally trained oncology models (group B) on a linear scale.}
\label{fig:3}
\end{figure}

OncoNote-RadBERT and OncoNoteBERT demonstrated substantially better fit to the held-out outpatient oncology corpus than either external zero-shot comparator, see Table \ref{tab:2} and Figure \ref{fig:3}. OncoNote-RadBERT achieved the strongest held-out language fit among the four encoders with validation loss reduced from 4.728 (for the zero-shot RadBERT baseline) to 0.7406, corresponding to a reduction in perplexity from 113.04 to 2.097. OncoNoteBERT achieved a validation loss of 1.041 and a perplexity of 2.832. PathologyBERT had the poorest fit to the outpatient oncology validation set, with a loss of 7.618 and perplexity of 2,035.03 when evaluated zero-shot using its native tokenizer. Comparison between OncoNote-RadBERT and RadBERT offers the clearest evidence of continued pretraining as they share the same tokenizer lineage (apart from the addition of \texttt{[REMOVED]} as a special token). In contrast, comparisons of OncoNote-RadBERT and OncoNoteBERT to PathologyBERT represent combined model-tokenizer fit because these models used different tokenizers.

\begin{table}[t]
\centering
\scriptsize
\caption{Encoder configuration and intrinsic masked language modelling performance}
\label{tab:2}
\begin{adjustbox}{max width=\linewidth}
\begin{tabular}{@{}p{0.18\linewidth}p{0.20\linewidth}p{0.20\linewidth}p{0.20\linewidth}p{0.18\linewidth}@{}}
\toprule
Model & Best validation loss & Perplexity & Best step & Actual steps \\
\midrule
Group A: ... &  &  &  &  \\

RadBERT & 4.728 & 113.04 & N/A & N/A \\

PathologyBERT & 7.618 & 2,035.03 & N/A & N/A \\

Group B: ... &  &  &  &  \\

OncoNote-RadBERT & 0.7406 & 2.097 & 86,000 & 88,000 \\

OncoNoteBERT & 1.041 & 2.832 & 196,000 & 200,000 \\
\bottomrule
\end{tabular}
\end{adjustbox}
\end{table}

\subsection{Tokenizer Efficiency}

\begin{figure}[t]
\centering
\includegraphics[width=0.92\linewidth]{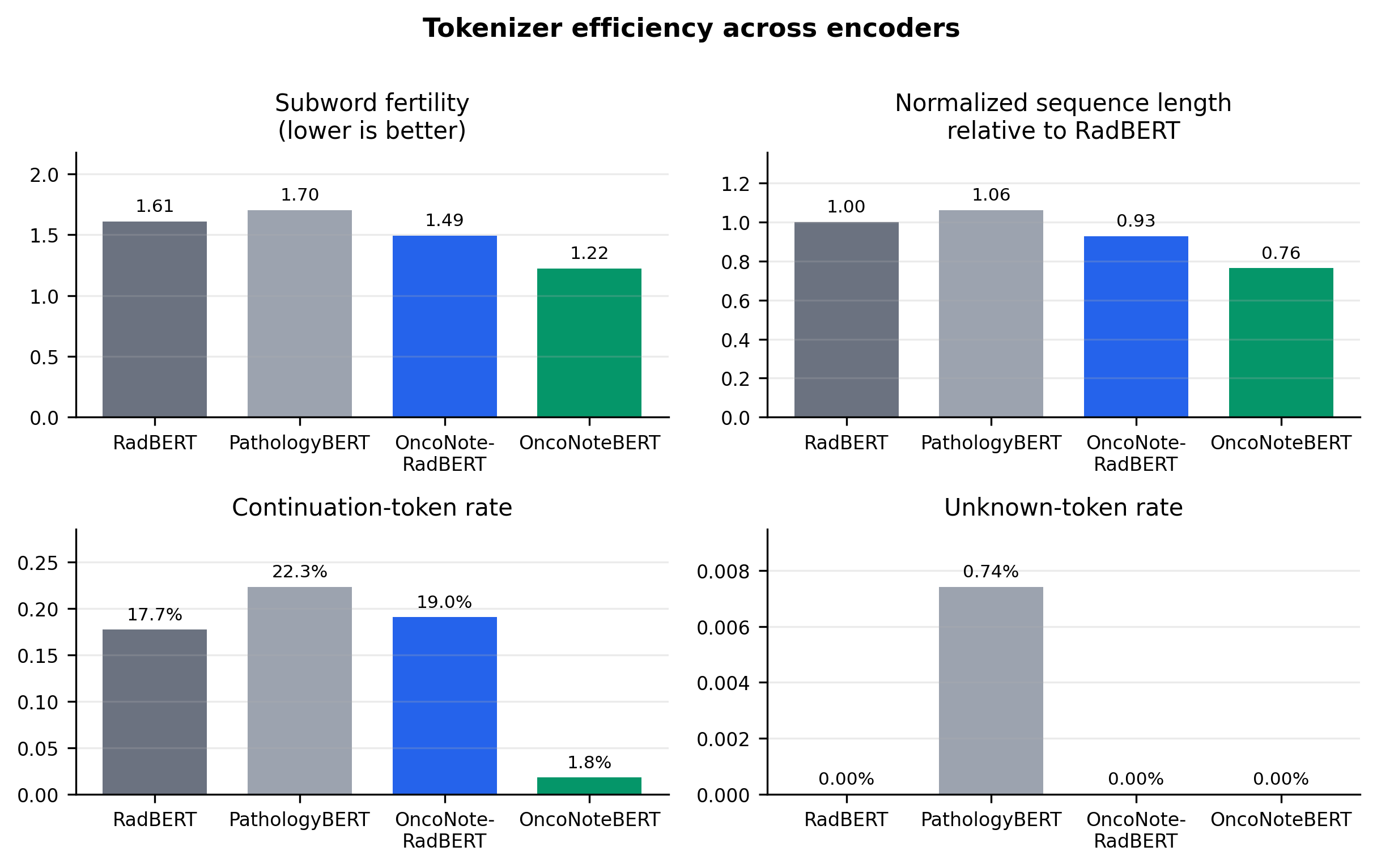}
\caption{Tokenizer efficiency across the four encoders. The figure compares subword fertility, continuation-token rate, unknown-token rate, and normalised sequence length on the shared document sample. For all these metrics, lower values indicate better performance.}
\label{fig:4}
\end{figure}

OncoNoteBERT demonstrated the greatest overall tokenizer efficiency on the 5,000-document sample (Figure \ref{fig:4}). Its mean subword fertility was 1.224 fragments per whitespace-delimited word, compared with 1.610 for RadBERT, 1.493 for OncoNote-RadBERT, and 1.702 for PathologyBERT. Its continuation-token rate was 0.01781, substantially lower than 0.1771 for RadBERT, 0.1905 for OncoNote-RadBERT, and 0.223 for PathologyBERT.

Unknown-token rates were negligible for RadBERT, OncoNote-RadBERT, and OncoNoteBERT, at 4.23e-8, 4.43e-8, and 5.59e-8, respectively, but were higher for PathologyBERT at 7.42e-3. OncoNoteBERT also had the shortest normalised sequence length, at 0.7636 relative to RadBERT, indicating that the same sampled documents required approximately 23.6\% fewer tokens under the oncology-specific tokenizer.

A targeted audit of oncology-related strings showed the practical effect of tokenizer design; see representative tokenisation outputs in Table \ref{tab:3}. OncoNoteBERT represented several specialised expressions as single tokens, including ECOG, T3N2M0, chemoradiotherapy, cisplatin, durvalumab, dysphagia, hypopharynx, mucositis, and pembrolizumab. In contrast, the inherited RadBERT tokenizer fragmented many of these terms into multiple subword pieces. Both OncoNote-RadBERT and OncoNoteBERT represented \texttt{[REMOVED]} as a single dedicated token, whereas RadBERT and PathologyBERT divided the placeholder into punctuation and lexical components.

\begin{table}[t]
\centering
\scriptsize
\caption{Tokenisation of selected oncology and de-identification expressions}
\label{tab:3}
\begin{adjustbox}{max width=\linewidth}
\begin{tabular}{@{}p{0.18\linewidth}p{0.20\linewidth}p{0.20\linewidth}p{0.20\linewidth}p{0.18\linewidth}@{}}
\toprule
Clinical string & RadBERT & PathologyBERT & OncoNote-RadBERT & OncoNoteBERT \\
\midrule
ECOG & e \#\#co \#\#g & ec \#\#og & e \#\#co \#\#g & ECOG \\

PD-L1 & p \#\#d - l \#\#1 & pd - l1 & p \#\#d - l \#\#1 & PD - L1 \\

T3N2M0 & t \#\#3 \#\#n \#\#2 \#\#m \#\#0 & t3 \#\#n \#\#2 \#\#m \#\#0 & t \#\#3 \#\#n \#\#2 \#\#m \#\#0 & T3N2M0 \\

\texttt{[REMOVED]} & [ removed ] & [ removed ] & \texttt{[REMOVED]} & \texttt{[REMOVED]} \\

best supportive care & best supportive care & best supportive care & best supportive care & best supportive care \\

chemoradiotherapy & ch \#\#em \#\#ora \#\#dio \#\#ther \#\#ap \#\#y & chemora \#\#di \#\#other \#\#apy & ch \#\#em \#\#ora \#\#dio \#\#ther \#\#ap \#\#y & chemoradiotherapy \\

cisplatin & c \#\#is \#\#p \#\#lat \#\#in & cis \#\#pla \#\#tin & c \#\#is \#\#p \#\#lat \#\#in & cisplatin \\

durvalumab & du \#\#rva \#\#lum \#\#ab & du \#\#r \#\#val \#\#uma \#\#b & du \#\#rva \#\#lum \#\#ab & durvalumab \\

dysphagia & d \#\#ys \#\#pha \#\#gia & dys \#\#pha \#\#gia & d \#\#ys \#\#pha \#\#gia & dysphagia \\

hypopharynx & h \#\#y \#\#pop \#\#har \#\#ynx & hypo \#\#pha \#\#ry \#\#nx & h \#\#y \#\#pop \#\#har \#\#ynx & hypopharynx \\

mucositis & m \#\#uc \#\#os \#\#itis & mucositis & m \#\#uc \#\#os \#\#itis & mucositis \\

pembrolizumab & p \#\#em \#\#bro \#\#li \#\#zu \#\#ma \#\#b & pem \#\#br \#\#oli \#\#z \#\#uma \#\#b & p \#\#em \#\#bro \#\#li \#\#zu \#\#ma \#\#b & pembrolizumab \\
\bottomrule
\end{tabular}
\end{adjustbox}
\end{table}

\subsection{Masked-Token Behaviour}

\begin{figure}[t]
\centering
\includegraphics[width=0.92\linewidth]{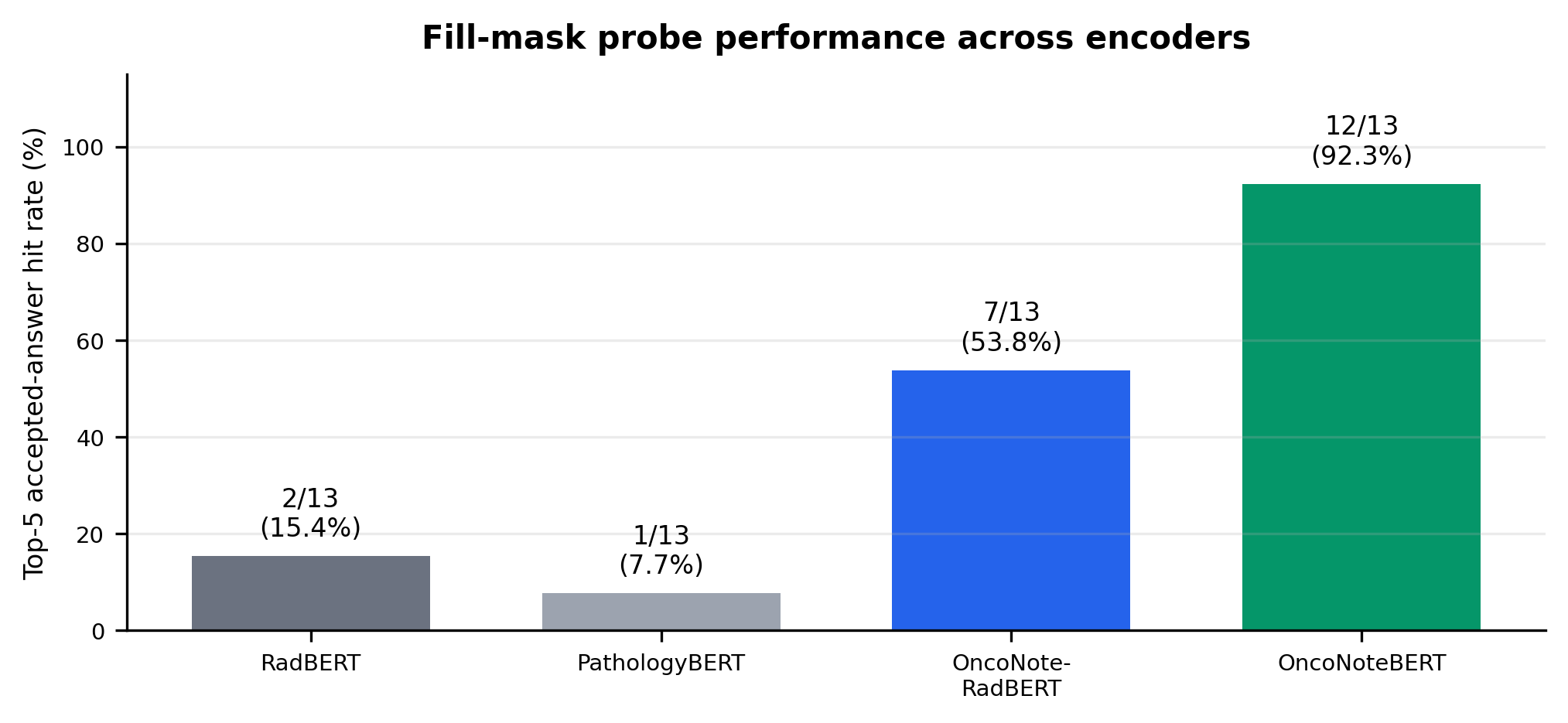}
\caption{Masked-token probe performance across encoders. The scorecard reports the number of prompts for which at least one prespecified clinically acceptable answer appeared among the five highest-ranked predictions. Higher values indicate better performance.}
\label{fig:5}
\end{figure}

OncoNoteBERT achieved the strongest performance on the 13 prespecified fill-mask prompts, returning an accepted answer among its five highest-ranked predictions for 12 prompts. OncoNote-RadBERT achieved a top-five accepted-answer result for seven prompts, compared with two for RadBERT and one for PathologyBERT (Figure \ref{fig:5}). The prompts and prespecified clinically acceptable answers are shown in Table \ref{tab:4}. This occurred despite OncoNote-RadBERT achieving lower validation perplexity, indicating that corpus-level MLM fit and targeted masked-token behaviour captured different aspects of model performance.

The stronger OncoNoteBERT result reflected both contextual prediction and tokenizer coverage because accepted expressions represented as single vocabulary tokens could be produced from one masked position. Because the probe set was small and based on prespecified acceptable answers, the results were treated as a qualitative audit rather than as a tokenizer-independent measure of clinical knowledge or downstream performance.

\begin{table}[t]
\centering
\scriptsize
\caption{Prespecified masked-token prompts and clinically acceptable answers}
\label{tab:4}
\begin{adjustbox}{max width=\linewidth}
\begin{tabular}{@{}p{0.18\linewidth}p{0.52\linewidth}p{0.24\linewidth}@{}}
\toprule
Category & Prompt & Accepted answers \\
\midrule
Privacy & Patient seen by Dr \texttt{[MASK]} in oncology clinic. & \texttt{[REMOVED]} \\

Privacy & He was seen on \texttt{[MASK]} for routine oncology follow-up. & \texttt{[REMOVED]} \\

Performance status & His \texttt{[MASK]} status deteriorated significantly. & performance \\

Lung cancer & Stage IV \texttt{[MASK]} cell lung cancer was discussed. & small; non; squamous; large \\

Lung cancer & PET scan demonstrated \texttt{[MASK]} avid mediastinal nodes. & fdg; FDG \\

Lung cancer & The patient started cisplatin and \texttt{[MASK]} chemotherapy. & pemetrexed; gemcitabine; etoposide \\

Head and neck cancer & The primary tumour involved the \texttt{[MASK]}. & oropharynx; larynx; nasopharynx; hypopharynx; tonsil \\

Head and neck cancer & He developed severe \texttt{[MASK]} after radiotherapy. & mucositis; dysphagia; xerostomia \\

Head and neck cancer & The tumour was \texttt{[MASK]} positive. & hpv; HPV; p16 \\

Treatment & Following chemoradiotherapy, the patient started consolidation \texttt{[MASK]}. & durvalumab; immunotherapy \\

Toxicity & Dose was reduced due to grade 3 \texttt{[MASK]}. & neutropenia; fatigue; diarrhoea; thrombocytopenia \\

Palliative care & Patient elected for best supportive \texttt{[MASK]} only. & care \\

Palliative care & Patient was referred to \texttt{[MASK]} care. & palliative; supportive \\
\bottomrule
\end{tabular}
\end{adjustbox}
\end{table}

The exploratory embedding audit indicated that continued oncology adaptation altered the RadBERT representation space while retaining substantial alignment with the original encoder. Across the 750 matched validation chunks, the mean cosine similarity between RadBERT and OncoNote-RadBERT representations was 0.7738. By contrast, representations produced by OncoNoteBERT and PathologyBERT had mean cosine similarities close to zero when compared directly with RadBERT: -0.0054 for RadBERT versus OncoNoteBERT and 0.0152 for RadBERT versus PathologyBERT. Note that these near-zero values should not be interpreted as evidence of poor representation quality because models trained from different random initialisations can encode similar information in different coordinate systems. The results therefore indicate that continued pretraining preserved substantially more of RadBERT's original structure, whereas the independently initialised models occupied different representation spaces. Summary statistics for the embedding audit are reported in Table \ref{tab:5}.

\begin{table}[t]
\centering
\scriptsize
\caption{Cosine similarity of matched validation representations relative to RadBERT}
\label{tab:5}
\begin{adjustbox}{max width=\linewidth}
\begin{tabular}{@{}llllll@{}}
\toprule
Encoder pair & Mean & SD & Minimum & Median & Maximum \\
\midrule
RadBERT vs OncoNote-RadBERT & 0.7738 & 0.0269 & 0.6624 & 0.7714 & 0.8701 \\

RadBERT vs OncoNoteBERT & -0.0054 & 0.0207 & -0.0572 & -0.0060 & 0.0544 \\

RadBERT vs PathologyBERT & 0.0152 & 0.0205 & -0.0742 & 0.0172 & 0.0682 \\
\bottomrule
\end{tabular}
\end{adjustbox}
\end{table}

\section{Discussion}

Representations learned from real-world outpatient oncology notes were not readily interchangeable with those from adjacent clinical domains such as radiology and pathology. RadBERT and PathologyBERT were relevant comparators but neither provided strong zero-shot model-tokenizer fit. Continued MLM pretraining from RadBERT substantially improved validation language fit when the target corpus differed in vocabulary, purpose, and longitudinal structure. Moreover, using a bespoke tokenizer in OncoNoteBERT resulted in the strongest tokenisation efficiency and the highest score on the targeted audit. The study was designed as a foundational model-introduction and characterisation study establishing the relative strengths of OncoNote-RadBERT and OncoNoteBERT before subsequent task-specific evaluation.

OncoNote-RadBERT achieved the lowest validation loss and perplexity, consistent with previous work showing that further pretraining on target-domain text can improve model performance \citep{gururangan2020dont}. It inherited representations from general BERT pretraining, biomedical adaptation in BioBERT, and radiology adaptation in RadBERT before continued pretraining on outpatient oncology notes. The comparison with RadBERT provides the clearest evidence of local adaptation because both models shared the same tokenizer lineage, apart from the addition of \texttt{[REMOVED]} as a special token. The reduction in perplexity from 113.04 to 2.097 therefore primarily reflects parameter adaptation, although the frequent \texttt{[REMOVED]} token may also have contributed. The PathologyBERT comparison is less controlled because it used a smaller, independent tokenizer; its high perplexity therefore reflects combined model-tokenizer fit rather than model-parameter quality alone.

The two local strategies improved different parts of the representation pipeline: continued pretraining produced the strongest corpus-level language fit, whereas from-scratch training with an oncology-native vocabulary reduced fragmentation and improved targeted clinical-token prediction. OncoNoteBERT required approximately 23.6\% fewer tokens than RadBERT and generated fewer chunks and fallback splits under the 512-token constraint.

Reduced fragmentation and shorter sequences provide representational and computational advantages, but they do not by themselves establish improved downstream performance. Biomedical language models may remain effective despite imperfect subword segmentation, and previous research has found that linguistically improved biomedical tokenisation does not necessarily produce corresponding gains across MLM and downstream tasks \citep{gutierrez2023biomedical}. The present findings therefore establish that OncoNoteBERT represents the outpatient oncology corpus more compactly, but the clinical value of this efficiency must be tested through downstream extraction and prediction tasks.

The fill-mask comparison was not independent of tokenizer design. Each prompt contained one masked position, so a model could return only one vocabulary token. An accepted expression represented as a single token by OncoNoteBERT could be predicted directly, whereas an expression divided into several WordPieces by RadBERT or PathologyBERT could not be produced in full from one masked position. The results therefore reflect both contextual prediction and vocabulary coverage; this interaction is informative in itself, but the findings should be read as a qualitative audit rather than a tokenizer-independent comparison of clinical knowledge. Future evaluation could score complete candidate strings using tokenizer-specific pseudo-log-likelihood or multiple-mask procedures. The handling of \texttt{[REMOVED]} illustrates the governance-sensitive nature of local clinical language modelling. De-identified notes often retain placeholders marking the presence and position of redacted information. By preserving \texttt{[REMOVED]} as a dedicated special token, the locally developed models could learn the linguistic contexts in which redacted content occurred without access to the underlying identifiable information. This does not constitute a formal privacy guarantee, nor does it show that the models cannot retain sensitive information present elsewhere in the corpus; rather, it offers a transparent way to represent an existing institutional de-identification convention within a data-resident workflow. A key limitation is that \texttt{[REMOVED]} does not distinguish among categories of redacted information. Where category-specific semantics matter, future models could use separate tokens such as \texttt{[NAME]}, \texttt{[DATE]}, or \texttt{[LOCATION]}, subject to de-identification system performance and governance requirements. This could preserve more grammatical and semantic information while continuing to exclude the identifiable value. The exploratory representation audit showed that OncoNote-RadBERT retained substantial alignment with RadBERT, as expected from their shared initialisation lineage. Near-zero cosine similarities between RadBERT and the independently initialised OncoNoteBERT and PathologyBERT models should not be read as evidence of inferior or unrelated representations, since independently trained networks may encode similar information in differently oriented coordinate systems. Direct cosine similarity is therefore most interpretable between models sharing an initialisation lineage. Future cross-model analyses may benefit from methods such as centred kernel alignment \citep{kornblith2019similarity}. Several limitations should be considered. The corpus came from a single UK specialist oncology centre and was restricted to lung and head-and-neck cancer notes, so external validity across institutions, tumour groups, and documentation practices remains to be established. Each locally developed model was trained once using one architecture and one prespecified configuration, so reported outcomes are point estimates rather than averages across initialisations or hyperparameter settings. Perplexity comparisons involving PathologyBERT and OncoNoteBERT reflect the combined effects of model parameters and tokenizer design and should not be treated as tokenizer-independent rankings. The fill-mask evaluation comprised only 13 prespecified prompts and did not capture the full range of clinically plausible completions, while the representation analyses were exploratory and did not establish model quality or clinical utility. The study focused on intrinsic representation evaluation rather than a specific downstream task. The findings therefore demonstrate language fit, tokenisation efficiency, contextual prediction behaviour, and representational differences, but do not establish improved clinical extraction or prediction. Subsequent work should evaluate both encoders on oncology tasks including treatment and toxicity extraction, treatment-response classification, phenotype identification, tumour-stage extraction, and survival or treatment-decision modelling.

\section{Conclusion}

We demonstrated that domain-specific training strategies can produce foundation representation models with substantially improved intrinsic performance and more efficient representation of specialised language within a governed, real-world oncology setting. OncoNote-RadBERT and OncoNoteBERT are intended to serve as reusable foundation representation backbones for subsequent task-specific studies within the approved clinical environment, although their downstream clinical value remains to be established through further evaluation.

\section*{Acknowledgments}

W.O. is supported by a Medical Research Council (MRC) PhD studentship
(grant reference 2897525) linked to the project ``Applying Natural Language
Processing to Real-World Patient Data to Optimise Cancer Care''
(MR/W007428/1), hosted at The University of Manchester.

\bibliographystyle{plainnat}
\bibliography{references}

\end{document}